\pdfoutput=1
\documentclass[11pt]{article}

\usepackage{acl}
\usepackage{times}
\usepackage{latexsym}
\usepackage[T1]{fontenc}
\usepackage[utf8]{inputenc}

\usepackage{microtype}
\usepackage{graphicx}

\usepackage{subcaption}

\title{Putting the \emph{Con} in Context:\\ Identifying Deceptive Actors in the Game of Mafia}

\author{Samee Ibraheem\thanks{\ \ Equal contribution.} , Gaoyue Zhou\footnotemark[1] , \and John DeNero \\
  University of California, Berkeley \\
  \texttt{\{sibraheem,zhougy\_99,denero\}@berkeley.edu} \\}

\begin{document}
\maketitle
\begin{abstract}

While neural networks demonstrate a remarkable ability to model linguistic content, capturing contextual information related to a speaker's conversational role is an open area of research. In this work, we analyze the effect of speaker role on language use through the game of Mafia, in which participants are assigned either an honest or a deceptive role. In addition to building a framework to collect a dataset of Mafia game records, we demonstrate that there are differences in the language produced by players with different roles. We confirm that classification models are able to rank deceptive players as more suspicious than honest ones based only on their use of language. Furthermore, we show that training models on two auxiliary tasks outperforms a standard BERT-based text classification approach. We also present methods for using our trained models to identify features that distinguish between player roles, which could be used to assist players during the Mafia game.

% Neural networks have allowed for a host of advances in Natural Language Processing, from text classification to language modeling. While these models demonstrate the ability to capture linguistic content, capturing contextual information in relation to one's conversational role is still an active area of research. This is especially relevant in online environments, where language may vary based on such attributes. In this work, we analyze the effect of role on language through the game of Mafia, in which participants may adopt either an honest or a deceptive role. In addition to building a framework to collect a dataset for the Mafia game, we demonstrate that there are differences in the language produced by players of each role and show that models are able to consistently rank deceptive parties as more suspicious than honest ones. We also present methods for using our trained models to identify features that distinguish between player roles and to assist players during the Mafia game. Our results have implications for real-life communication paradigms, providing possible avenues for hiding aspects of one's identity or discovering aspects of another's.

% This document is a supplement to the general instructions for *ACL authors. It contains instructions for using the \LaTeX{} style files for ACL conferences. 
% The document itself conforms to its own specifications, and is therefore an example of what your manuscript should look like.
% These instructions should be used both for papers submitted for review and for final versions of accepted papers.

\end{abstract}

\section{Introduction}

% Language is a powerful tool that enables not only for the communication of content between interlocutors, but also context. For instance, if one were to say that they are a foreigner, this communicates something about their status in relation to their current location, which would not be the case if they were instead in their country of origin. Though languages vary in the degree and manner to which such attributes may be communicated, this is a common thread among them, and thus one which we would expect our Natural Language Processing systems to be equipped to handle.

Correct interpretation of language must take into account not only the meaning of utterances, but also characteristics of the speaker and the context in which their utterances are produced. Modeling the impact of this context on language is still challenging for NLP systems. For example, differences in language identification accuracy, speech recognition word error rates, and translation quality have been observed on the basis of attributes such as a speaker's gender, race, dialect, or role \citep{blodgett, tatman2017effects, tatman2017gender, stanovsky}. Moreover, these systems systematically underperform on data generated by those in the minority, having implications for the ethics and fairness of using these technologies.

This work explores language used for deception: a type of speaker context that is particularly challenging to model because it is intentionally hidden by the speaker. To do so, we collect and release a set of records for the game of Mafia, in which each player is assigned either an honest or a deceptive role. Then, we develop models that distinguish players' roles based only on the text of the players' dialog. We describe two auxiliary tasks that improve classification accuracy over a BERT-based text classifier.

The novel contributions of this paper include:
\begin{enumerate}
\item A methodology for collecting records of online Mafia games and a dataset collected from 460 human subjects,
\item Three classification models that can distinguish between honest and deceptive players,
\item An approach for identifying features of the game dialog text that can be used to help identify deceptive players during the game.
\end{enumerate}
The task of identifying deception in dialog is far from solved. Our classification methods, while not accurate enough to reliably identify deceptive players in a game, do show that the text of a dialog in the setting we study does contain information about the roles of the participants, even when those participants are motivated to hide those characteristics by deceiving the listener. Although the models and results described in this work only apply to a particular game setting rather than dialog in general, the approaches we describe are general in character and therefore may inform future work on determining speaker roles from the contents of dialog.

\section{Background \& Related Work}

The game of Mafia is particularly well-suited for the goal of determining whether the deceptive participants in a conversation can be identified from the contents of their utterances.

\subsection{Deception in Language}

% Humans are a largely collaborative species: strangers on the internet exchange goods and services, online social groups form and thrive, people work remotely and communicate on online platforms. However, people sometimes have goals that incentivize them to deceive others. People may bend the truth during negotiations, and in an online environment, people may not know what conversations are with deceptive or malicious entities without the aid of normal audiovisual cues. Understanding what cues and interaction styles people adopt when behaving deceptively, or seeking to detect deceptive behavior, will be crucial to both developing automated detection and a greater understanding of the complex interactions that people use in deception and revelation. Previous work indicates that people struggle with telling apart lies from truth, especially with deceptive statements \citep{bond}. This raises the question of what strategies deceptive actors use to avoid detection, as well as what strategies honest actors use to discover deceivers.

Humans are a largely collaborative species. However, people sometimes have goals that incentivize them to deceive others. Understanding what cues and interaction styles people adopt when behaving deceptively will be crucial to both developing automated detection and a greater understanding of the complex interactions that people use in deception and revelation. Previous work indicates that people struggle with telling apart lies from truth, especially with deceptive statements \citep{bond}. This raises the question of what strategies deceptive actors use to avoid detection, as well as what strategies honest actors use to discover deceivers.

% Moreover, there is a distinction between the following:

% \begin{itemize}
%     \item a falsehood: a statement that is not true
%     \item a lie: a statement that the speaker does not believe
%     \item deception: the act of convincing another person to hold a false belief
% \end{itemize}

Deception is a difficult topic to study, however, because of its inherent complexity: multiple people with different motivations are trying to evaluate one another, while contending with group obligations and accusations, over a period of time that involves planning, taking actions, and responding to others' actions. Moreover, there is a distinction between a falsehood, which is a statement that is not true, a lie, which is a statement that the speaker does not believe, and deception, which is the act of convincing another person to hold a false belief. Whereas falsehoods and lies are properties of statements, deceptive intent is a characteristic of the speaker. Therefore, though deceptive speakers may tell falsehoods and lies, they might also provide truthful statements, and vice versa for honest speakers, thus rendering the truth conditions of individual utterances as unreliable indicators of deception. We are interested in how people solve these dual problems of deceiving and detecting deception, which requires a paradigm wherein we can observe all agents' actions and communication while simultaneously knowing agents' underlying incentives and goals. We thus turn to a game with a rich history of deception research: Mafia.

Previous work on detecting deception from linguistic cues has explored scenarios that either mimic or are taken directly from real-world investigations of potentially deceptive actors. \citet{derrick} showed that deceptive parties take longer to formulate responses and use fewer words in the context of chat-based communication. \citet{burgoon} similarly found that deceivers sent briefer chat messages. \citet{fuller} demonstrated the effectiveness of training classifiers to identify deceptive language in relation to crimes, and found that word quantity was a particularly useful feature. \citet{fornaciari} also found surface-level features useful in detecting deceptive statements in a criminal context, specifically through the investigation of Italian court documents, while \citet{mihalcea} found that written lies were easier to detect than transcripts of spoken ones. \citet{abouelenien} took a multimodal approach to deception detection, finding that non-contact approaches were able to match or exceed the performance of those that were more invasive.

\subsection{The Game of Mafia}

Researchers have also examined deception in games, focusing on settings such as Diplomacy or negotiation over a set of items \citep{lewis, niculae}. In addition, there has been some work exploring the effects of biased voting on group decision making \citep{kearns}. The game of Mafia specifically has attracted attention, and researchers have analyzed data from various online game communities. \citet{zhou2008} discovered differences between deception across cultural communities by analyzing data from an online Chinese Mafia game, \citet{pak2011social} used social network analysis to detect deceivers using the epicmafia.com website, and \citet{deruiter} collected and trained models on a dataset from the online Mafiascum forum. Researchers have also studied the game of Werewolf, a variant of Mafia. \citet{chittaranjan} used audio information to classify deceptive parties, while \citet{demyanov} used video information. \citet{braverman} and \citet{migdal} developed a mathematical model of the Mafia game, assuming that all votes are cast at random, which allowed them to analyze how mafia and bystander win rates varied with role distribution in a highly controlled version of the game. \citet{bi} showed that under certain conditions, the strategy of mafia pretending to be bystanders is suboptimal.

Most of the deception-oriented games that have been studied in the natural language processing literature provided individual incentives to the players. Mafia allows for the study of patterns of deception that arise when incentives are only at the group level. In addition, whereas using datasets of online Mafia games presents a rich source of deceptive language, the complicated rule sets of games on these forums makes it challenging to isolate specific strategies that participants use to engage in and detect deceptive behavior. In contrast to work using video or audio, we assume that players do not have access to any audiovisual clues about others' roles in order to focus on the role of language. This work takes these factors into account by studying a controlled environment that nonetheless supports the use of complex strategies for deceiving and detecting deceptive behavior.

\section{Dataset}

\begin{figure*}
 \centering
\begin{subfigure}{.5\textwidth}
\centering
\includegraphics[scale = .25, trim={0cm 0cm 0cm 0cm},clip]{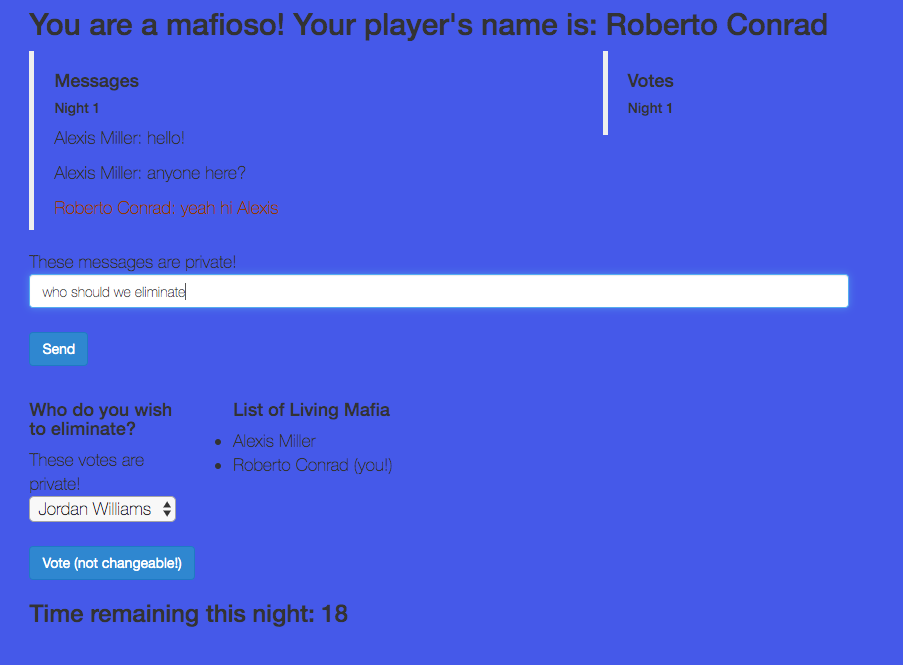}
\end{subfigure}%
\begin{subfigure}{.5\textwidth}
\centering
\includegraphics[scale = .237, trim={0cm 0cm 0cm 0cm},clip]{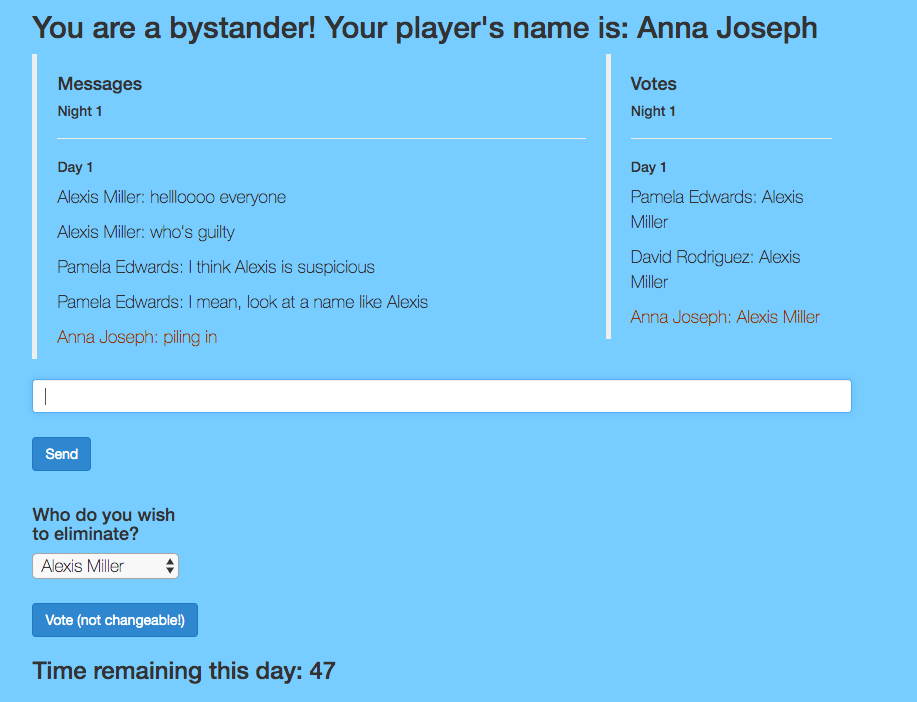}
\end{subfigure}
\caption{Mafia experiment screenshot during \textbf{(left)} first nighttime phase, with participant as a mafioso, and \textbf{(right)} first daytime phase, with participant as a bystander (note that mafia messages are not visible to the bystander).}
\label{fig:mafia}
\end{figure*}

A total of 460 English-speaking participants based in the United States were recruited from Amazon Mechanical Turk using the experiment platform Dallinger\footnote{\url{http://github.com/dallinger/Dallinger}}. Between 4 and 10 participants were recruited for each Mafia game: 1 to 2 participants were designated mafia, and the rest were bystanders. Forty-four of these Mafia games are included in the final analysis. Participants were paid \$2.50 for completing the task, plus bonuses for time spent waiting for other participants to arrive in a chatroom to begin the experiment. Waiting was paid at \$5/hour.

Upon recruitment, participants were shown a consent form, per IRB approval, followed by an instructional video and accompanying transcript describing how to play the text-based Mafia game using an interface we developed (see Appendix). After they completed a quiz demonstrating they understood the information, they entered a waiting room until the desired number of participants was reached. Participants were then assigned a role (mafioso or bystander) and fake name, after which they began playing the game.

The game dynamics were as follows. Each mafia member was aware of the roles of their fellow mafia members and thus, by process of elimination, knew the roles of the bystanders. However, the bystanders did not know the true role of anyone else in the game. The goal of the mafia was to eliminate bystanders until the number of mafia was greater than or equal to that of the bystanders. The goal of the bystanders was to identify and eliminate all of the mafia members. Since the incentive structure was set up such that bystanders benefited from true beliefs about who the mafia members were, whereas mafia members benefited from false beliefs, bystanders were thus motivated to be honest actors, whereas mafia members were motivated to be deceptive actors in the Mafia game. The game proceeded in phases, alternating between nighttime and daytime (Figure~\ref{fig:mafia}). During the nighttime, mafia members could secretly communicate to decide on who to eliminate, after which they discretely voted, and the person with the majority vote was eliminated from the game. If there was a tie, one of the people involved in the tie was randomly chosen to be eliminated. During the daytime, everyone was made aware of who was eliminated during the nighttime, and then all players could openly communicate to decide who to eliminate. All the players then voted publicly, and the person with the majority vote was eliminated and announced to be a bystander or mafioso. Thus, during the nighttime mafia could secretly communicate and eliminate anyone, whereas during the daytime mafia could participate in the voting and communication protocols in the same way as bystanders. The game proceeded until there was a winning faction according to the goals described above.

\begin{figure*}[!ht]
\centering
\includegraphics[width=0.8\textwidth]{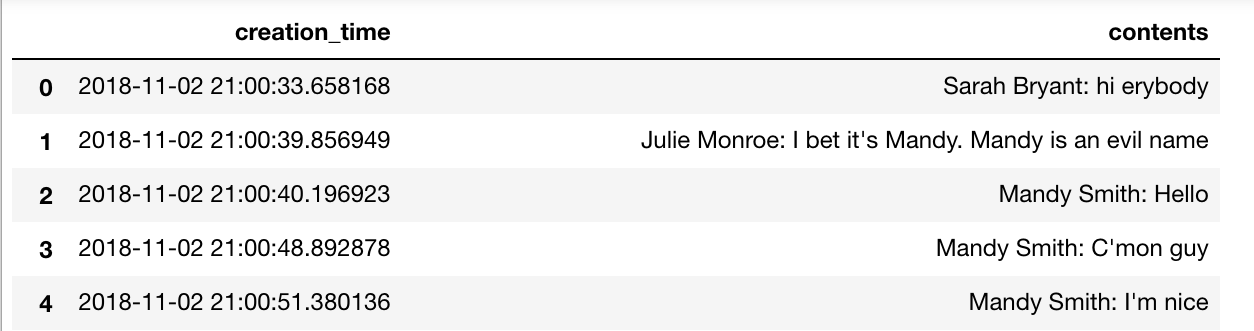}
\caption{Example messages (utterances) in a game. \textit{creation\_time} is the time at which the message was sent. The \textit{contents} consists of the name of the sender, as well as the message, separated by a colon and space.}
\label{fig:text_data}
\end{figure*}

From these experiments, we collected a dataset consisting of both mafia and bystander utterances over the course of each game, as well as the participants' voting behavior. Dataset statistics appear in Table~\ref{tab:dataset_stats}. Figure~\ref{fig:text_data} displays a snippet of the daytime dialog from one Mafia game. As shown, many utterances are either social interactions (eg. "hi erybody") or discussions about what to do in the game, such as accusations or comments about voting (eg. "I bet it's Mandy...").

\begin{table}
\centering
\begin{tabular}{ccccc}
\hline
 & \textbf{M} & \textbf{B} & \textbf{T} & \\
\hline
\textbf{Total \#players} &  87  &	334 &	421 \\
\textbf{Avg \#players per game} & 1.98 &	7.59 &	9.57 \\
\textbf{Std \#players per game} & 0.15 &	1.21 &	1.28 \\
\textbf{Total \#utt} & 770 &	1392 &	2162 \\
\textbf{Avg \#utt per game} & 17.5 &	31.64 &	49.14 \\
\textbf{Std \#utt per game} & 10.45	& 17.2 &	24.44 \\
\textbf{Total \#players w/ utt} & 84 &	265	& 349 \\
\textbf{Perc players w/o utt} & 0.042 &	0.958 &	1 \\
\hline
\end{tabular}
\caption{Dataset statistics. \emph{\#} is short for \emph{number of}. \textbf{M} and \textbf{B} denote the mafioso and bystander classes, respectively, while \textbf{T} denotes the total number for both groups. The last row shows the distribution of roles among the players with no utterances throughout the game. Note that nearly all of the no-utterance players are bystanders.}
\label{tab:dataset_stats}
\end{table}

Upon further inspection of the data, we can observe several strategies used by mafia members to deceive bystanders:

\begin{enumerate}
    \item Mafia members may suggest that there is not enough information to decide on who to eliminate, despite their knowledge of everyone's roles (eg. ``Should we wait to eliminate someone?'' / ``It's a little early to tell.'' / ``It's a shot in the dark.''),
    \item Mafia members may raise suspicion about another player, despite knowing that said player is a bystander (eg. hmm ok analyzing this conversation....I think bianca was a little to flippant in how she was like "sucks to be andrew" haha / I'm going to vote bianca. she's so casual with life and death),
    % \item Mafia members may take advantage of bystanders forgetting information (eg. suggesting someone to be eliminated despite having already killed that person during the nighttime),
    % \item Mafia members may deliberately vote for a player during the nighttime in order to frame another player during the daytime, despite knowing that player is a bystander
    \item Mafia members may invent a false motive and assign that motive to another player, despite knowing that the player is a bystander (eg. It might be Jonathan Kim... killing off Erin who accused him "yesterday").
\end{enumerate}

% \section{Task}

\section{Approach}

Given our mafia dataset, there are several tasks that one might address, for example, predicting participants' daytime voting behavior or generating mafia members' nighttime dialog. As our aim is to identify deceptive actors, however, we focus on predicting participants' roles, i.e. bystander or mafioso. Due to the asymmetry in the knowledge available to each group and the goals which incentivize bystanders to increase true belief and mafia members to reduce it, the bystanders are said to take on an honest role in the game, whereas the mafia members take on a deceptive role. To focus on the relationship between language and deception, we ignore voting behavior and consider just the daytime dialog in the game, as only the mafia members were able to converse during the nighttime. As shown in Table~\ref{tab:dataset_stats}, since most of the players with no utterances are bystanders, we only consider players who make at least one utterance throughout the game.

To investigate whether linguistic information can be used to identify players' roles, we train and evaluate classifiers that predict the role of a particular player. Since we have a small dataset, we chose to fine-tune pre-trained Transformer models rather than train them from scratch \citep{vaswani}. To predict the role for a player $p$, we construct an input representation $r(C, p)$ of the full game dialog $C$ that encodes the player of interest $p$. We develop three approaches which differ in both the dialog representation function $r$ and the modeling approach.

\subsection{Standard Classification}

Our baseline approach uses a standard BERT-based text classifier \citep{devlin}. To classify player $p$ via the full record of the game $C$, let boolean variable $M_p$ be true if $p$ is a mafioso. Let $T_p$ be the concatenation\footnote{Utterances are concatenated with an end-of-sentence delimiter after each utterance.} of utterances made by $p$. We train BERT parameters $\theta_M$ to predict $P(M_p|T_p; \theta_M)$. 

% While this model can be used directly to predict the role of a player, we find that weighting this prediction by the prior probability $P(M_p)$, which is the fraction of players who are mafia members across all training games, improves performance:

% $$
%     P(M_p|T_p) = \frac{P(M_p)P(M_p|T_p;\theta_M)}{\sum_{R\in\{M,\neg M\}} P(R_p)P(R_p|T_p;\theta_M)}
%     % \label{eq:std_bert}
% $$

% where $P(\neg M_p) = 1 - P(M_p)$.

This approach, which provides as input to the classifier only the utterances of the player to be classified, outperformed an alternative representation $r(C, p)$ that included the entire record of all utterances by all players.

% Our baseline approach uses a standard BERT-based text classifier \citep{devlin}. The best performing method of representing the game sets $r(t, p)$ to the concatenation\footnote{Utterances are concatenated such that there is an end-of-sentence delimiter between them.} of all of the utterances made by player $p$ throughout the game, where the goal is to predict the role of $p$ based on their speech. We define the following variables:

% \begin{itemize}
% \item $M$: a boolean denoting whether the speaker is a mafioso
% \item $T$: the text of all of the target player's utterances
% \item $\theta$: the BERT model parameters
% \end{itemize}
% Our goal is to model the conditional probability $P(M|T)$ for all players in a game.

% We fine-tune a BERT model, which gives us $P(M|T, \theta)$ and $P(\neg M|T, \theta)$.
% We found that performance could be improved by incorporating the prior probability that a player is a mafioso $P(M)$, which is the fraction of players with a deceptive role. We incorporate this prior by scaling the BERT model predictions:

% \begin{equation}
%     P(M|T) = \frac{P(M)P(M|T,\theta)}{\sum_{R\in\{M,\neg M\}} P(R)P(R|T,\theta)}
%     \label{eq:std_bert}
% \end{equation}
% where $P(\neg M) = 1 - P(M)$.

%We also explored using the speech of all players as input to the BERT model, as well as using Random Forest and Bag of Words methods to classify players from just their own speech, but found that using BERT with only the speech of the target player outperformed these alternatives, which only have slight improvements than random predictions.

\subsection{Auxiliary Tasks}

Limiting the input representation $r$ to contain only the speech of the player $p$ being classified is not ideal; correctly interpreting a dialog requires considering all other players' statements as well. We introduce two auxiliary tasks that involve the entire game dialog $C$:

\begin{enumerate}
\item Given all of the prior utterances, is a bystander or a mafia member more likely to have produced the current utterance? (\emph{Utterance Classification})
\item Given all of the prior utterances, what current utterance would a player produce, given that they are a bystander or a mafia member? (\emph{Utterance Generation})
\end{enumerate}

We develop a BERT-based classification model for task~1 and fine-tune the GPT-2 language model for task~2 \citep{radford}. Then, we use each of these auxiliary models to classify the role of a particular player $p$ in the game.

%The features used for the previous approaches are a drastic over-simplification of what happens in the Mafia game, and the model fails to make use of information about other players in the same game, even though these interactions between players are crucial to the game setting. To address this problem, we consider an auxiliary problem: given the context of prior game utterances, what information can be used to predict the current user's role? Specifically, we investigate the following: (1) Given all of the prior utterances, is a bystander or a mafia member more likely to have produced the current utterance? (2) Given all of the prior utterances, what would be the current utterance of a bystander or a mafia member? We utilize BERT for Question 1 and GPT-2 for Question 2, leveraging its decoding capability \citep{radford}.

\subsubsection{Utterance Classification}

To classify player $p$ using the auxiliary task of utterance classification, let boolean variable $S_i$ be true if utterance $C_i$ was made by a mafioso (rather than a bystander). Let $C$ be the full record of utterances in the game and $C_{\leq i}$ be the concatenation of all utterances $C_1 \dots C_i$.  We train BERT parameters $\theta_S$ to predict $P(S_i | C_{\leq i}; \theta_S)$. Finally, let $I_p$ be the set of indices of utterances by player $p$. $M$ relates to $S$ in that if $M_p$ is true, then $S_i$ is true for all $i \in I_p$. We thus calculate

$$P(M_p|C; \theta_S) \propto \frac{\sum_{i \in I_p} P(S_i | C_{\leq i}; \theta_S)}{N},$$

\noindent where $N = |I_p|$.

\subsubsection{Utterance Generation}

To classify player $p$ using the auxiliary task of utterance generation, we fine-tune GPT-2 to generate utterance $C_i$ conditioned on prior utterances $C_{<i}$ and the role $S_i$ of the speaker that produced $C_i$. From Bayes' rule, we have $P(M_p|C) \propto P(M_p) P(C|M_p).$ To estimate $P(C|M_p)$, let $C_p$ include all $C_i$ for $i \in I_p$. We make the simplifying assumption that $P(C|M_p) \propto P(C_p|M_p)$, which assumes that the utterances made by players other than $p$ are independent of the role of player $p$. Then, if $M_p$ is true, $S_i$ is true for all $i \in I_p$, and so,

$$P(C_p|M_p; \theta_C) = \prod_{i \in I_p} P(C_i | C_{<i}, S_i; \theta_C).$$

Using the full dialog $C$, the final probability of player $p$ being mafioso is calculated as follows:
% We introduce a hyperparameter $\beta$ that adjusts the relative influence of the prior and prediction:\footnote{$\beta$ was set to 4.1 for our experiments, which was chosen to maximize accuracy. Again, this hyperparameter does not affect the ranking of different players by their probability of being mafia, which is the primary evaluation metric described in Section~\ref{metrics}.}
% \begin{equation}
%     P(M_p|C) = \frac{P(M_p)}{P(M_p) + P(\neg M_p)\left(\frac{P(C_p|\neg M_p; \theta_C)}{P(C_p|M_p; \theta_C)}\right)^\beta}
%     \label{eq:gpt2_2}
% \end{equation}

\begin{equation}
    P(M_p|C) = \frac{P(M_p)P(C_p|M_p; \theta_C)}{\sum_{R\in\{M,\neg M\}}
    P(R_p)P(C_p|R_p; \theta_C)}
    \label{eq:gpt2_2}
\end{equation}

% $$
%     P(M_p|T_p) = \frac{P(M_p)P(M_p|T_p;\theta_M)}{\sum_{R\in\{M,\neg M\}} P(R_p)P(R_p|T_p;\theta_M)}
%     % \label{eq:std_bert}
% $$

% We define the following variables:
% \begin{itemize}
% \item $M$: a boolean denoting whether the speaker is a mafioso
% \item $C$: the text of all prior game utterances
% \item $U$: the current utterance
% \end{itemize}
% Our goal is to model the conditional probability $P(M|C, U)$ for all (player, utterance) pairs in a game.

% We fine-tune a GPT-2 model to answer Question 2, which gives us $P(U|C, M)$ and $P(U|C, \neg M)$. To make predictions about players' roles, we calculate the probability $P(M|C, U)$, again according to Bayes' rule: 
% \begin{equation}
%     P(M|C, U) = \frac{P(M|C)P(U|C, M)}{\sum_{R\in\{M,\neg M\}}P(R|C)P(U|C, R)}
%     \label{eq:gpt2_1}
% \end{equation}
% where $P(U|C, M)$ and $P(U|C, \neg M)$ are given by the GPT-2 model, aggregating over all of the utterances for player $p$ by summing their losses. $P(M|C)$ and $P(\neg M|C)$ can again be simplified to $P(M)$ and $P(\neg M)$, respectively.

% To take into account the potential noise and inaccuracy in our trained language model, we further introduce a hyperparameter $\alpha$\footnote{$\alpha$ was set to 4.1 for our experiments.} to adjust the relative importance between the prior distribution and our language model, so that the probability $P(M|C, U)$ becomes:
% \begin{equation}
%     P(M|C, U) = \frac{P(M)}{P(M) + P(\neg M)\left(\frac{P(U|C, \neg M)}{P(U|C, M)}\right)^\alpha} 
%     \label{eq:gpt2_2}
% \end{equation}

\begin{figure}[t]
\begin{center}
\includegraphics[width=1\linewidth]{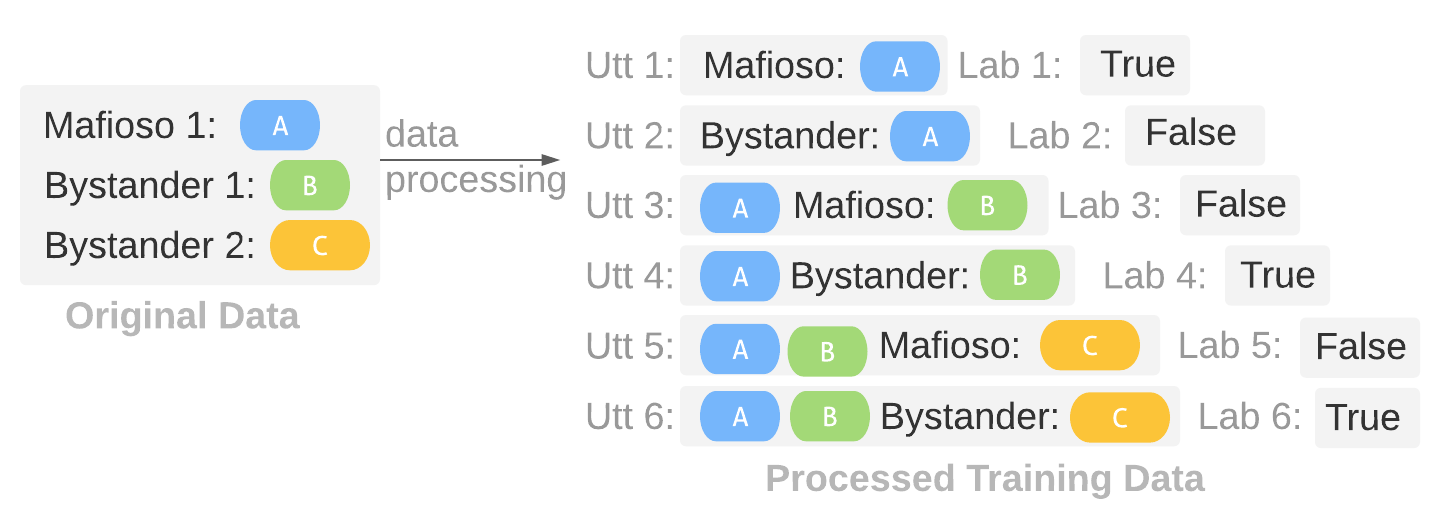}
\end{center}
  \caption{Data processing for fine-tuning BERT. The original data is shown on the left-hand side, while the right-hand side shows the processed data containing two versions of each utterance, one assuming that the target player is a mafioso and one assuming that they are a bystander, with the prior conversation context preceding each and labels corresponding to whether the assumed role matches the actual role of the player.}
\label{fig:bert_data}
\end{figure}

\begin{figure}[t]
\begin{center}
\includegraphics[width=1\linewidth]{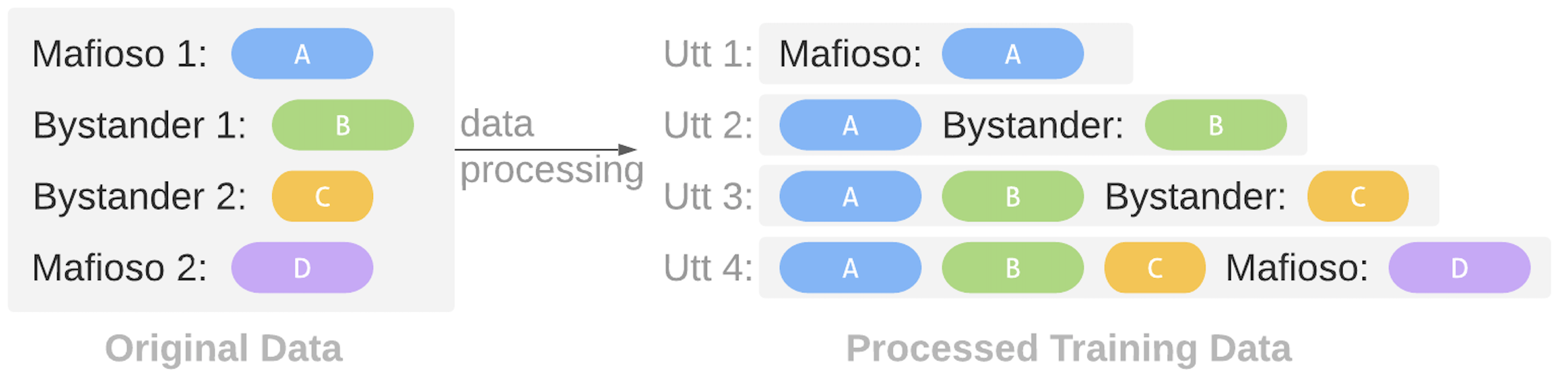}
\end{center}
  \caption{Data processing for fine-tuning GPT-2. The original data is shown on the left-hand side, while the right-hand side shows the processed data containing a version of the corresponding utterance with the prior conversation context preceding.}
\label{fig:gpt2_data}
\end{figure}

\subsection{Data Processing}

To train models for utterance classification (using BERT) and utterance generation (using GPT-2), we perform data processing procedures on the games' original dataset to create input representations $r(C, p)$ for each player $p$ and obtain our training datasets as shown in Figures~\ref{fig:bert_data} and \ref{fig:gpt2_data}. The left side of each figure shows a snippet of a game's data, where "Mafioso" and "Bystander" denote the true roles of the players. The utterances to the right of each figure are training examples used for fine-tuning the BERT and GPT-2 models. Structuring the data in this way provides both the prior context of utterances and the current utterance that happened within this context. This not only gives us the information needed for the auxiliary tasks, but also provides us with more training examples, as we only have 44 games and only 421 players in total, with only 2162 total utterances. Moreover, this mimics the real game scenario from the bystander view in that they can only confirm their own role, but no one else's, which is the appropriate setting for us in which to detect deception.

\begin{figure}[t]
\begin{center}
\includegraphics[width=1\linewidth]{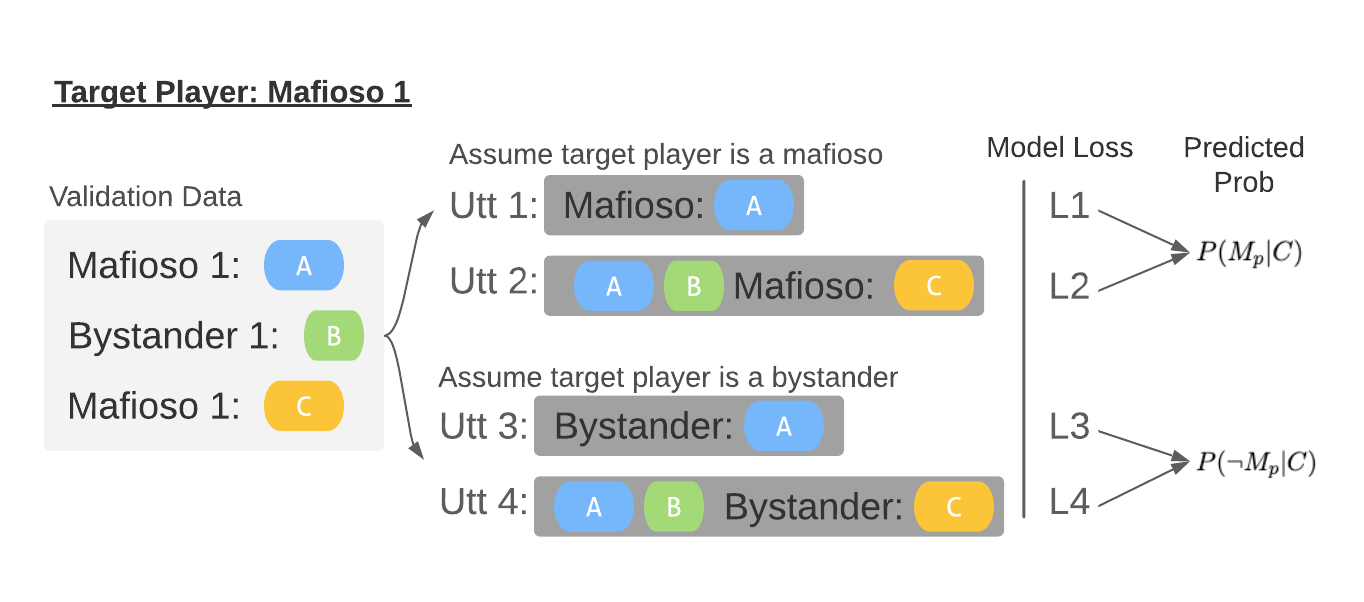}
\end{center}
  \caption{Prediction pipeline for our fine-tuned GPT-2 model. Similar to the pipeline used to produce the training utterances, for prediction, there are now two versions of each, one assuming that the target player is a mafioso and one assuming that they are a bystander. The losses for each utterance of the target player are summed together in order to calculate the mafia and bystander probabilities as described in Equation~\ref{eq:gpt2_2}.}
\label{fig:gpt2_pred}
\end{figure}

Figure~\ref{fig:gpt2_pred} shows the pipeline for using the GPT-2 model to predict players' roles. Let us assume that the target player for whom we want to predict their role is Mafioso 1. From the original game log on the left, we first perform the data processing scheme from Figure~\ref{fig:gpt2_data} twice, assuming that the target player is a mafioso (top of Figure~\ref{fig:gpt2_pred}) and a bystander (bottom of Figure~\ref{fig:gpt2_pred}). Using our trained GPT-2 model, we then obtain a loss for each utterance denoted by L1 through L4. Summing all the losses for each role, as they denote log probabilities, we calculate $P(M_p|C)$ and $P(\neg M_p|C)$ via Equation~\ref{eq:gpt2_2}. The target player's role as predicted by the model is finally given by comparing the two probabilities. A similar process is used to calculate $P(M_p|C)$ and $P(\neg M_p| C)$ for the utterance classification BERT model.

\section{Experiments}

We train three fine-tuned models on the corpus of Mafia game records and compare their performance to a random baseline. The specifications for the baseline and models can be found below, and the results are shown in Table~\ref{tab:res}.

% \begin{table*}
% \centering
% \begin{tabular}{cccccccc}
% \hline
%  & \textbf{Avg Rank} & \textbf{Avg Rank/Game} & \textbf{Accuracy} & \textbf{Maf Prec} & \textbf{Maf Rec} & \textbf{Bys Prec} & \textbf{Bys Rec}\\
% \hline
% \textbf{Random}    & 19.0 & 3.4 & 0.62 & 0.26 & 0.26 & 0.74 & 0.74 \\
% % \textbf{BoW}    & 22.3 & 4.1 & 0.59 & 0.25 & 0.3 & 0.74 & 0.69 \\
% % \textbf{RF}     & 17.5 & 3.3 & 0.64 & 0.33 & 0.4 & 0.78 & 0.72 \\
% \textbf{Std Class} & 20.9 & 3.6 & \textbf{0.69} & 0.33 & 0.20 & 0.75 & \textbf{0.86} \\
% \textbf{Utt Class} & 16.6 & 2.9 & 0.67 & \textbf{0.40} & 0.60 & 0.83 & 0.69 \\
% \textbf{Utt Gen}   & \textbf{11.4} & \textbf{2.1} & 0.64 & \textbf{0.40} & \textbf{0.80} & \textbf{0.89} & 0.59 \\
% \hline
% \end{tabular}
% \caption{Experiment results on the validation set for random baseline (\textbf{Random}), standard classification (\textbf{Std Class}), utterance classification (\textbf{Utt Class}), and utterance generation (\textbf{Utt Gen}) approaches. Utterance generation outperforms all other methods in terms of average ranking overall and per game while also maintaining high accuracy, recall, and precision.}
% \label{tab:res}
% \end{table*}
\begin{table*}
\centering
\begin{tabular}{cccccccc}
\hline
 & \textbf{Avg Rank} & \textbf{Avg Rank/Game} & \textbf{Accuracy} & \textbf{Maf F1-score} & \textbf{Bys F1-score}\\
\hline
\textbf{Random}    & 19.0 & 3.4 & 0.62 & 0.26 & 0.74 \\
% \textbf{BoW}    & 22.3 & 4.1 & 0.59 & 0.25 & 0.3 & 0.74 & 0.69 \\
% \textbf{RF}     & 17.5 & 3.3 & 0.64 & 0.33 & 0.4 & 0.78 & 0.72 \\
% \textbf{Std Class} & 20.9 & 3.6 & 0.69 & 0.25 & 0.80\\
\textbf{Std Class} & 17.9 & 3.0 & 0.69 & 0.4 & 0.79\\
% \textbf{Utt Class} & 16.6 & 2.9 & 0.67 &  0.48 & 0.75\\
\textbf{Utt Class} & 14.5 & \textbf{1.8} & \textbf{0.74} & \textbf{0.50} & \textbf{0.83}\\
\textbf{Utt Gen}   & \textbf{11.4} & 2.0 & \textbf{0.74} & \textbf{0.50} & \textbf{0.83}\\
\hline
\end{tabular}
\caption{Experiment results on the validation set for random baseline (\textbf{Random}), standard classification (\textbf{Std Class}), utterance classification (\textbf{Utt Class}), and utterance generation (\textbf{Utt Gen}) approaches. Methods that use auxiliary tasks (\textbf{Utt Class} and \textbf{Utt Gen}) outperform other methods in terms of average ranking overall and per game while also maintaining higher accuracy and F1-score for each class.}
\label{tab:res}
\end{table*}

% \subsection{Baselines}

% \textbf{Random Classifier (Random)}

\subsection{Random Baseline}

This random classifier classifies each player as a mafioso or a bystander with probabilities equal to the prior distribution of each class, estimated as the ratio of roles across all training games. This serves as a baseline to be compared to for all other methods. In the game setting, this mimics a bystander player with only public information of how many mafia and bystanders are in the game.

\subsection{Standard Classification}

We initialize the model by loading a pre-trained BERT Base model (12 layers, 768 hidden dimension size, 12 attention heads). We train with a maximum sequence length of 256, which is sufficient for our post-processed dataset, setting the batch size to 16, the learning rate to 1e-5, and the maximum number of epochs to 25.

% \medskip
%  \noindent \textbf{Bag of Words (BoW)}  We compare against the Bag of Words method, which takes into account only the composition of each player's utterances to see if this would be enough to differentiate between roles. We put all utterances of each player into a paragraph, replace the references to other players' names with special words to prevent the model from learning game-dependent features, then embed each player into a feature vector with entries denoting the counts of words in the dictionary. Finally, we perform logistic regression on the resulting feature vectors.
 
% \medskip
%  \noindent \textbf{Random Forest (RF)} In order to determine whether surface-level features would be enough to distinguish between players, we also compare against a random forest classifier. To train this classifier, we first featurized our data using the following: how often a participant sent a daytime text, the average number of texts they sent per round, and the average length in characters of a participant's texts. We then used these features as input to the random forest classifier.

% \subsection{Auxiliary Model Specifications}
% \textbf{BERT}

\subsection{Utterance Classification}

We initialize the model by loading a pre-trained BERT Base model (12 layers, 768 hidden dimension size, 12 attention heads). We train with a maximum sequence length of 512, which is sufficient for our post-processed dataset, setting the batch size to 5, the learning rate to 5e-5, and the maximum number of epochs to 25.

% \medskip
%  \noindent \textbf{GPT-2}

\subsection{Utterance Generation}

We initialize the model by loading a pre-trained 12-layer GPT-2 model with an embedding size of 768. For the dataset, we set the maximum length of each sentence to be 512, which is sufficient for our dataset after post-processing. During training, we set the batch size to be 5 and the learning rate to be 1e-5. We train the model for a maximum of 100 epochs.

% We train each model for a maximum of 100 epochs, although it usually converges within 40 epochs.
 
\subsection{Metrics} \label{metrics}

%Since our ultimate goal is to detect deception and use our model to potentially help players to distinguish mafia players during the game, we are most concerned with the model's performance on mafia players relative to bystanders in the game. Therefore, besides the accuracy for this binary classification problem, we also consider metrics that better satisfy this requirement. Since players need to decide which player to vote for in each round, intuitively, assigning a probability-of-being-mafia to every player and then ranking all players based on this probability would be particularly helpful for players to make rational decisions.

These approaches each estimate a probability $P(M_p|C)$ that a player $p$ is a mafioso given the full record of game texts $C$. In Mafia, bystanders do not declare who is and is not a mafioso, but instead vote each day to eliminate one of the players. Because the act of voting involves choosing one player among them all, a natural metric for evaluating the usefulness of a model is to order all players $p$ from greatest to least $P(M_p|C)$, their probability of being a mafioso under the model, and then to compute the average rank of the true mafia members. Therefore, the first metric in Table~\ref{tab:res} is the average ranking of all mafia members when each player is ranked by $P(M_p|C)$ across the entire validation set composed of 5 games. It is also natural to consider player ranking within a single game, so we calculate the average ranking of mafia members within each game as a second metric. Smaller average ranking for mafia members means that the model is able to assign mafia players a high $P(M_p|C)$ relative to bystanders, which is desired.

% These rank-based metrics are not affected by the scaling terms $\alpha$ in Equation~\ref{eq:bert_2} and $\beta$ in Equation~\ref{eq:gpt2_2}.

In addition, we evaluate the accuracy of the classifiers and the F1-score for each class. To calculate these metrics, we first assign the mafioso label to the top $k$ players with the highest $P(M_p|C)$ and the rest of the players with the bystander label, where $k$ is the known number of mafia among all validation games ($k=10$ in our case). Aside from the ranking metrics, these give further information of the models' quality after utilizing available game information. 

\subsection{Results and Analysis}

We trained all models on 39 training games and evaluated on the remaining 5 validation games. The evaluation results are shown in Table~\ref{tab:res}. We have a total of 49 players in the validation games, but only considered the 39 players who had spoken at least one utterance throughout the game when calculating the metrics. Players with no utterances are almost exclusively bystanders and are therefore easy to classify without considering language.

% We see that our fine-tuned GPT-2 model achieves significantly lower average mafia ranking and average mafia ranking per game compared to the other approaches, and it achieves the best mafia precision and recall. Note that in our case where we hope to detect deception and help players to vote out mafia players, we are more concerned with the ranking of mafia members than the overall accuracy, since the classification error does not measure the usefulness of the model to make decisions in the context of the game. Although the overall accuracy for our GPT-2 model is not prominent, from a player's perspective, it still gives the most accurate suggestions on who to vote for.

First, we see that it is possible to achieve an average rank that is smaller than the random baseline, which demonstrates that there is information in the dialog about the roles of players, despite the fact that mafia members seek to hide their role while conversing. However, standard classification is comparable to random. Next, we observe that both models using auxiliary tasks outperform the standard classifier in rank-based metrics, which demonstrates that the auxiliary tasks provide useful inductive bias for the mafia classification task. Additionally, the accuracy is similar for all approaches, including random classification, which indicates that there is not enough information in the text of a Mafia game for these models to determine players' roles reliably. If the goal of the game were to guess the role of each player individually, then always guessing bystander (i.e. the majority class) would be the best strategy. However, since the goal for the bystanders is to vote to eliminate a mafia member each round, the utterance classification and utterance generation approaches, which achieve the lowest average mafia ranking per game and overall, respectively, are the most favorable.

Note that the precision for the mafia is much lower than that of the bystanders for all models. This is due to the usual lack of information available to predict that any player is a mafioso, which makes finding the mafia a much harder task than finding bystanders.

% TODO: why gpt2 outperforms bert in terms of ranking

\section{Discussion}

The decoding ability of the GPT-2 model provides us a more straightforward way to understand what the model has learned. Given a prompt sentence, we can use our fine-tuned GPT-2 model to generate what a mafioso and a bystander would say. A few examples are shown in Table~\ref{tab:gen}. From these examples, we inspect the following features that the model might be capturing to distinguish between mafia and bystanders: Feature 1: Referring to other players. Feature 2: Expressing confusion. Feature 3: Referring to others for elimination purposes. Feature 4: Asking for suggestions on who to eliminate.

\begin{table}
\centering
\begin{tabular}{p{1.9cm}|l}
\hline
\textbf{Prompt} & \textbf{Generated Utterance} \\
\hline
lets kill P1. & \begin{tabular}{@{}l@{}}
                   \textbf{M:} sorry P1 :(\\
                   \textbf{B:} hello all\\
                 \end{tabular} \\
\hline
who thinks P3 is Mafia? & \begin{tabular}{@{}l@{}}
                   \textbf{M:} No i'm a bystander\\
                   \textbf{B:} No idea\\
                 \end{tabular} \\
\hline
That sounds suspicious... & \begin{tabular}{@{}l@{}}
                   \textbf{M:} P6 is mafia\\
                   \textbf{B:} Why yall want to eliminate me?\\
                 \end{tabular}\\
\hline
hi team. Hello!. Hi. & \begin{tabular}{@{}l@{}}
                   \textbf{M:} Who is the mob person?\\
                   \textbf{B:} hello\\
                 \end{tabular}\\
\hline
\end{tabular}
\caption{Utterances generated by our GPT-2 model given different prompts. \textbf{M} and \textbf{B} are shorthand for Mafioso and Bystander respectively, and P1, P3, and P6 denote the names of other players in the game.}
\label{tab:gen}
\end{table}

To confirm that our fine-tuned GPT-2 model captures some of the above features, we hand-label these features on 5 training games and 1 validation game, obtain each player's feature vector, and see whether there exists a correlation between the model's predicted $P(M_p|C)$ for validation players and the similarity of their feature vectors compared to the training set mafioso and bystander players. These feature vectors are shown in Table~\ref{tab:train_hand}, where each entry denotes the average number of features per player of each role. As an example, for the first column, each mafioso player says 2 utterances having Feature 1 throughout the game on average, while each bystander player says 1.06 utterances having Feature 1 on average. We define the first row as a vector $v_1$ and the second row as $v_2$ for future references.

\begin{table}
\centering
\begin{tabular}{ccccc}
\hline
 & \textbf{Feat 1} & \textbf{Feat 2} & \textbf{Feat 3} & \textbf{Feat 4} \\
\hline
\textbf{Mafioso}   & 2.00 & 0.00 & 1.30 & 0.40 \\
\textbf{Bystander} & 1.06 & 0.27 & 0.65 & 0.10 \\
\hline
\end{tabular}
\caption{The average count per role for each of four hand-labeled features (number of references to other players, level of confusion, number of references to other players for elimination, and number of requests for who to eliminate) as identified by our GPT-2 model on 5 training games.}
\label{tab:train_hand}
\end{table}

Table~\ref{tab:val_hand} shows the hand-labeled feature vectors for all 10 players in a validation game (first 4 columns, F1 to F4) ranked by the model's predicted $P(M_p|C)$. We define a metric function $D(u) = \| u - v_1 \|^2 -  \| u - v_2 \|^2$ for a validation player's feature vector $u$. The smaller $D(u)$ is, the closer $u$ is to $v_1$ than $v_2$, and hence the more mafia-like they are with respect to players in the training games. We can see that for players of higher rank, their $D(u)$ are negative with larger magnitudes. Referring to the true labels in the rightmost column ($M$ for Mafioso and $B$ for Bystander), the first row also explains how our model can fail to predict the true role of some players: even though this player is a bystander, they act more like the mafia than other bystanders according to these hand-labeled features because they are regularly referencing and accusing other players.

\begin{table}
\centering
\begin{tabular}{cccccccc}
\hline
 & \textbf{F1} & \textbf{F2} & \textbf{F3} & \textbf{F4} & \textbf{D(u)} & \textbf{Pred} & \textbf{Truth} \\
\hline
\textbf{P0} & 4 & 0 & 2 & 0 & -5.9  & 0.98 & B \\
\textbf{P1} & 2 & 0 & 2 & 0 & -2.1  & 0.93 & M \\
\textbf{P2} & 5 & 0 & 5 & 0 & -11.7 & 0.78 & M \\
\textbf{P3} & 2 & 0 & 2 & 0 & -2.1  & 0.63 & B \\
\textbf{P4} & 4 & 2 & 1 & 1 & -4.1  & 0.47 & B \\
\textbf{P5} & 3 & 0 & 2 & 0 & -4.0  & 0.43 & B \\
\textbf{P6} & 0 & 0 & 0 & 0 & 4.2   & 0.42 & B \\
\textbf{P7} & 1 & 0 & 1 & 0 & 1.0   & 0.40 & B \\
\textbf{P8} & 0 & 0 & 0 & 0 & 4.2   & 0.00 & B \\
\textbf{P9} & 0 & 0 & 0 & 0 & 4.2   & 0.00 & B \\
\hline
\end{tabular}
\caption{Features of each player (P0 to P9) in a validation game. For each row, F1 to F4 give the feature vector $u$ for the respective player. $D(u)$ gives the similarity of $u$ compared to the training feature vectors $v_1$ and $v_2$. Players are sorted by $Pred$, the probability $P(M_p|C)$ given by our GPT-2 model, and $Truth$ gives the true label for each player ($M$ for Mafioso, $B$ for Bystander). Since P8 and P9 have no utterances throughout the game, as per our heuristic, they are predicted to be bystanders with $P(M_p|C) = 0$.}
\label{tab:val_hand}
\end{table}

\section{Limitations \& Potential Risks}

We find that we are able to train models to help differentiate players with different roles in the game of Mafia based only on their language use, as well as to identify features that may distinguish between these roles. We also noticed that the mafia were twice as likely to win the Mafia game than were the bystanders. These findings lead us to believe that the bystanders may benefit from being provided hints based on our model's predictions and identified features. However, there are several ethical considerations in regards to using these methods. First, as our model is trained on this particular version of mafia, the specific models trained would not apply to other cases of deceptive language use. Applying these models to out-of-domain data, or even adapting this general approach to new settings,  may yield unexpected results. Our experimental results only establish the effectiveness of our approach on the game of Mafia. Future work must evaluate these approaches on other deception detection tasks before they can be safely deployed in real-world scenarios. Next, information that may aid bystanders in detecting deception may also aid mafia members in being deceptive. Though mafia members may attempt to use it for this purpose, because our model is trained to increase true belief, which is directly in line with the bystander goal to identify the truth and against the mafia goal to obscure it, our approach is inherently more useful to bystanders. However, since the models we evaluate are far from perfectly accurate, there is a risk that users using these models for hints would rely too much on their output and thereby be misled. More work should be done to increase the model's performance in order to mitigate this risk.

\section{Conclusion}

%\section{Conclusion & Future Work}

% We find that we are able to train models to differentiate players with different roles in the game of Mafia based only on their language use, as well as to identify features that may distinguish between these roles. We also noticed that the mafia were twice as likely to win the Mafia game than were the bystanders. These findings lead us to believe that the bystanders may benefit from being provided suggestions for whom to eliminate given our model's predictions and identified features. However, information that may aid bystanders may also aid mafia members in their deception.

% We thus hope to run further experiments within our framework in order to compare the behavior of bystanders with such guidance to those of bystanders under our original setting.

How one uses language depends not only on the content they wish to convey, but also on the context within which they convey it, and speaker attributes such as conversational role contribute to such context. In this work, we leveraged an environment for which roles are explicitly labelled in order to make progress toward the task of deception detection, an essential task to protect users in our increasingly virtual world.

% \nocite{Ando2005,borschinger-johnson-2011-particle,andrew2007scalable,rasooli-tetrault-2015,goodman-etal-2016-noise,harper-2014-learning}

\section*{Acknowledgements}

We would like to thank Vael Gates, Aida Nematzadeh, Tom Griffiths, and the Dallinger team for their invaluable help in the development of the Mafia experiment. This material is based upon work supported by the National Science Foundation Graduate Research Fellowship Program under Grant No. DGE
1752814.

\bibliography{anthology,custom}

\begin{thebibliography}{25}
\expandafter\ifx\csname natexlab\endcsname\relax\def\natexlab#1{#1}\fi

\bibitem[{Abouelenien et~al.(2014)Abouelenien, P{\'e}rez-Rosas, Mihalcea, and
  Burzo}]{abouelenien}
Mohamed Abouelenien, Veronica P{\'e}rez-Rosas, Rada Mihalcea, and Mihai Burzo.
  2014.
\newblock Deception detection using a multimodal approach.
\newblock In \emph{{Proceedings of the 16th International Conference on
  Multimodal Interaction}}, pages 58--65.

\bibitem[{Bi and Tanaka(2016)}]{bi}
Xiaoheng Bi and Tetsuro Tanaka. 2016.
\newblock Human-side strategies in the {Werewolf} game against the stealth
  werewolf strategy.
\newblock In \emph{{International Conference on Computers and Games}}, pages
  93--102. Springer.

\bibitem[{Blodgett and O'Connor(2017)}]{blodgett}
Su~Lin Blodgett and Brendan O'Connor. 2017.
\newblock Racial disparity in natural language processing: A case study of
  social media {African-American English}.
\newblock \emph{arXiv preprint arXiv:1707.00061}.

\bibitem[{Bond~Jr and DePaulo(2006)}]{bond}
Charles~F Bond~Jr and Bella~M DePaulo. 2006.
\newblock Accuracy of deception judgments.
\newblock \emph{Personality and Social Psychology Review}, 10(3):214--234.

\bibitem[{Braverman et~al.(2008)Braverman, Etesami, and Mossel}]{braverman}
Mark Braverman, Omid Etesami, and Elchanan Mossel. 2008.
\newblock {Mafia: A theoretical study of players and coalitions in a partial
  information environment.}
\newblock \emph{The Annals of Applied Probability}, 18(3):825--846.

\bibitem[{Burgoon et~al.(2003)Burgoon, Blair, Qin, and Nunamaker}]{burgoon}
Judee~K Burgoon, J~Pete Blair, Tiantian Qin, and Jay~F Nunamaker. 2003.
\newblock Detecting deception through linguistic analysis.
\newblock In \emph{{International Conference on Intelligence and Security
  Informatics}}, pages 91--101. Springer.

\bibitem[{Chittaranjan and Hung(2010)}]{chittaranjan}
Gokul Chittaranjan and Hayley Hung. 2010.
\newblock Are you a werewolf? detecting deceptive roles and outcomes in a
  conversational role-playing game.
\newblock In \emph{{2010 IEEE International Conference on Acoustics, Speech and
  Signal Processing (ICASSP)}}.

\bibitem[{de~Ruiter and Kachergis(2018)}]{deruiter}
Bob de~Ruiter and George Kachergis. 2018.
\newblock The {Mafiascum} dataset: A large text corpus for deception detection.
\newblock \emph{arXiv preprint arXiv:1811.07851}.

\bibitem[{Demyanov et~al.(2015)Demyanov, Bailey, Ramamohanarao, and
  Leckie}]{demyanov}
Sergey Demyanov, James Bailey, Kotagiri Ramamohanarao, and Christopher Leckie.
  2015.
\newblock Detection of deception in the {Mafia} party game.
\newblock In \emph{{Proceedings of the 2015 ACM on International Conference on
  Multimodal Interaction}}, pages 335--342.

\bibitem[{Derrick et~al.(2013)Derrick, Meservy, Jenkins, Burgoon, and
  Nunamaker~Jr}]{derrick}
Douglas~C Derrick, Thomas~O Meservy, Jeffrey~L Jenkins, Judee~K Burgoon, and
  Jay~F Nunamaker~Jr. 2013.
\newblock Detecting deceptive chat-based communication using typing behavior
  and message cues.
\newblock \emph{ACM Transactions on Management Information Systems (TMIS)},
  4(2):1--21.

\bibitem[{Devlin et~al.(2018)Devlin, Chang, Lee, and Toutanova}]{devlin}
Jacob Devlin, Ming-Wei Chang, Kenton Lee, and Kristina Toutanova. 2018.
\newblock Bert: Pre-training of deep bidirectional transformers for language
  understanding.
\newblock \emph{arXiv preprint arXiv:1810.04805}.

\bibitem[{Fornaciari and Poesio(2013)}]{fornaciari}
Tommaso Fornaciari and Massimo Poesio. 2013.
\newblock Automatic deception detection in {Italian} court cases.
\newblock \emph{Artificial Intelligence and Law}, 21(3):303--340.

\bibitem[{Fuller et~al.(2011)Fuller, Biros, and Delen}]{fuller}
Christie~M Fuller, David~P Biros, and Dursun Delen. 2011.
\newblock An investigation of data and text mining methods for real world
  deception detection.
\newblock \emph{Expert Systems with Applications}, 38(7):8392--8398.

\bibitem[{Kearns et~al.(2009)Kearns, Judd, Tan, and Wortman}]{kearns}
Michael Kearns, Stephen Judd, Jinsong Tan, and Jennifer Wortman. 2009.
\newblock Behavioral experiments on biased voting in networks.
\newblock In \emph{{Proceedings of the National Academy of Sciences 106.5}},
  pages 1347--1352.

\bibitem[{Lewis et~al.(2017)Lewis, Yarats, Dauphin, Parikh, and Batra}]{lewis}
Mike Lewis, Denis Yarats, Yann~N. Dauphin, Devi Parikh, and Dhruv Batra. 2017.
\newblock {Deal or no deal? End-to-end learning for negotiation dialogues.}
\newblock \emph{arXiv preprint arXiv:1706.05125}.

\bibitem[{Migda\l{}(2010)}]{migdal}
Piotr Migda\l{}. 2010.
\newblock {A mathematical model of the {Mafia} game.}
\newblock \emph{arXiv preprint arXiv:1009.1031}.

\bibitem[{Mihalcea et~al.(2013)Mihalcea, P{\'e}rez-Rosas, and Burzo}]{mihalcea}
Rada Mihalcea, Ver{\'o}nica P{\'e}rez-Rosas, and Mihai Burzo. 2013.
\newblock Automatic detection of deceit in verbal communication.
\newblock In \emph{{Proceedings of the 15th ACM on International Conference on
  Multimodal Interaction}}, pages 131--134.

\bibitem[{Niculae et~al.(2015)Niculae, Kumar, Boyd-Graber, and
  Danescu-Niculescu-Mizil}]{niculae}
Vlad Niculae, Srijan Kumar, Jordan Boyd-Graber, and Cristian
  Danescu-Niculescu-Mizil. 2015.
\newblock {Linguistic harbingers of betrayal: A case study on an online
  strategy game.}
\newblock \emph{arXiv preprint arXiv:1506.04744}.

\bibitem[{Pak and Zhou(2011)}]{pak2011social}
Jinie Pak and Lina Zhou. 2011.
\newblock A social network based analysis of deceptive communication in online
  chat.
\newblock In \emph{{Workshop on E-Business}}, pages 55--65. Springer.

\bibitem[{Radford et~al.(2019)Radford, Wu, Child, Luan, Amodei, Sutskever
  et~al.}]{radford}
Alec Radford, Jeffrey Wu, Rewon Child, David Luan, Dario Amodei, Ilya
  Sutskever, et~al. 2019.
\newblock Language models are unsupervised multitask learners.
\newblock \emph{OpenAI blog}, 1(8):9.

\bibitem[{Stanovsky et~al.(2019)Stanovsky, Smith, and Zettlemoyer}]{stanovsky}
Gabriel Stanovsky, Noah~A Smith, and Luke Zettlemoyer. 2019.
\newblock Evaluating gender bias in machine translation.
\newblock \emph{arXiv preprint arXiv:1906.00591}.

\bibitem[{Tatman(2017)}]{tatman2017gender}
Rachael Tatman. 2017.
\newblock Gender and dialect bias in youtube’s automatic captions.
\newblock In \emph{Proceedings of the First ACL Workshop on Ethics in Natural
  Language Processing}, pages 53--59.

\bibitem[{Tatman and Kasten(2017)}]{tatman2017effects}
Rachael Tatman and Conner Kasten. 2017.
\newblock Effects of talker dialect, gender \& race on accuracy of bing speech
  and youtube automatic captions.
\newblock In \emph{INTERSPEECH}, pages 934--938.

\bibitem[{Vaswani et~al.(2017)Vaswani, Shazeer, Parmar, Uszkoreit, Jones,
  Gomez, Kaiser, and Polosukhin}]{vaswani}
Ashish Vaswani, Noam Shazeer, Niki Parmar, Jakob Uszkoreit, Llion Jones,
  Aidan~N Gomez, {\L}ukasz Kaiser, and Illia Polosukhin. 2017.
\newblock Attention is all you need.
\newblock In \emph{Advances in neural information processing systems}, pages
  5998--6008.

\bibitem[{Zhou and Sung(2008)}]{zhou2008}
Lina Zhou and Yu-wei Sung. 2008.
\newblock Cues to deception in online chinese groups.
\newblock In \emph{{Proceedings of the 41st Annual Hawaii International
  Conference on System Sciences (HICSS 2008)}}, pages 146--146. IEEE.

\end{thebibliography}
\bibliographystyle{acl_natbib}

% \pagebreak

% \newpage

\appendix

\section{Mafia Instructions}

Below is a transcript of the instructions that were provided to participants before playing the Mafia game in our experiments:

"In this experiment, you will play a version of the party game "Mafia".
You are going to play the game of Mafia (also known as Werewolf) with other participants. You are either part of the mafia (a mafioso) or a bystander. The mafia will know who is in the mafia, but the bystanders will not. There will always initially be more bystanders than mafia. There will be one or more mafia members.
The goal of the mafia is to eliminate the bystanders one by one until the mafia are equal in number to them. The goal of the bystanders is to correctly guess the identity of the mafia and eliminate them all before the mafia win.
There are two phases to this game, nighttime and daytime; at the end of each, a participant is eliminated from the game:
\begin{enumerate}
    \item In the \textbf{nighttime} phase, only the mafia can converse and decide who they want to eliminate. Specifically, if you are a mafioso, you will talk in a chatroom, then use a drop-down menu to select who you want to remove. Mafia will have 1 minute to do this. If there is more than one mafioso and the mafia disagree about who to eliminate, one of the mafia's choices will be selected randomly. If you are a bystander, you will wait out this time, as you are sleeping during the night.
    
    \item Everyone is awake during the \textbf{daytime} phase. The participant who was eliminated during the night will be announced: if you were eliminated, you will be sent to the end of the game and compensated. The remaining participants will converse (for 2 minutes and 30 seconds) and decide who to eliminate, where the goal of the bystanders is to eliminate a member of the mafia, and the goal of the mafia is to eliminate a bystander. By the end of this time, everyone needs to select a name from the drop-down menu. (If there are multiple mafia, the mafia will be reminded of each others' names in separate text on this page.) The participant with the most votes will be eliminated, except in the case of a tie, in which a randomly-selected vote will be eliminated. The eliminated participant and their identity (bystander or mafia) will be announced, and that participant will be sent to the end of the game and compensated.
\end{enumerate}
The game will continue, alternating between nighttime and daytime, until either all of the mafia are removed (\textit{bystanders win!}) or there are equal numbers of mafia and bystanders (\textit{mafia win!})"

\end{document}